\documentclass[11pt,a4paper]{article}

\usepackage[utf8]{inputenc}
\usepackage[T1]{fontenc}
\usepackage{tipa}
\usepackage{microtype}
\usepackage{hyperref}
\usepackage{url}
\usepackage{booktabs}
\usepackage{multirow}
\usepackage{graphicx}
\usepackage{amsmath}
\usepackage{natbib}
\usepackage{xcolor}

\definecolor{darkblue}{rgb}{0, 0, 0.5}
\hypersetup{colorlinks=true, citecolor=darkblue, linkcolor=darkblue, urlcolor=darkblue}

\title{When Does Data Augmentation Help? Evaluating LLM and
Back-Translation Methods for Hausa and Fongbe NLP}

\author{
  Mahounan Pericles Adjovi$^1$ \quad Roald Eiselen$^2$ \quad Prasenjit Mitra$^1$ \\[0.5em]
  $^1$Carnegie Mellon University Africa, Kigali, Rwanda \\
  $^2$Centre for Text Technology, North-West University, Potchefstroom, South Africa \\[0.3em]
  \texttt{madjovi@andrew.cmu.edu} \quad
  \texttt{Roald.Eiselen@nwu.ac.za} \quad
  \texttt{prasenjm@andrew.cmu.edu}
}

\date{}

\begin{document}

\maketitle

\begin{abstract}
Data scarcity limits NLP development for low-resource African languages.
Across the MasakhaNER 2.0 and MasakhaPOS benchmarks, available training data
ranges from a few hundred to several thousand sentences per language, making
augmentation strategies an attractive remedy. We evaluate two data augmentation
methods---LLM-based generation (Gemini 2.5 Flash) and back-translation (NLLB-200)---for
Hausa and Fongbe, two West African languages that differ substantially in LLM
generation quality as established by prior evaluations~\citep{robinson2023chatgpt,hendy2023good}.
We assess augmentation on named entity recognition (NER) and part-of-speech (POS)
tagging using MasakhaNER 2.0 and MasakhaPOS benchmarks.

Our results reveal that augmentation effectiveness depends on \emph{task type}
rather than language or LLM quality alone. For NER, neither method improves over
baseline for either language; specifically, LLM augmentation reduces Hausa NER by
0.24\% F1 and Fongbe NER by 1.81\% F1. For POS tagging, LLM augmentation improves
Fongbe by 0.33\% accuracy while back-translation improves Hausa by 0.17\%;
back-translation reduces Fongbe POS by 0.35\% and has negligible effect on Hausa POS.
Strikingly, the same LLM-generated synthetic data produces opposite effects across
tasks for Fongbe---hurting NER while helping POS---suggesting task structure governs
augmentation outcomes more than synthetic data quality. Human inspection of generated
examples, discussed in Section~\ref{sec:discussion}, reveals systematic label-quality
differences across tasks that explain this divergence.

These findings challenge the assumption that LLM generation quality predicts
augmentation success. Practitioners should evaluate augmentation impact per task
rather than assuming benefits transfer across tasks. Beyond African NLP, our findings
provide actionable guidance: data augmentation should be treated as a task-specific
intervention rather than a universally beneficial preprocessing step.
\end{abstract}

\section{Introduction}
\label{sec:introduction}

Natural language processing for low-resource African languages faces a persistent
data scarcity challenge~\citep{nekoto-etal-2020-participatory,joshi2020state}.
Limited training data constrains the performance of sequence-labeling models,
reduces generalization, and makes it difficult to build reliable NLP pipelines for
tasks such as named entity recognition (NER) and part-of-speech (POS) tagging.
Across the Masakhane benchmarks, available training data varies widely:
MasakhaNER 2.0 ranges from roughly 1,000 to 8,000 sentences across 20 African
languages~\citep{adelani2022masakhaner}, while MasakhaPOS covers fewer than 1,000
sentences for many languages~\citep{dione2023masakhapos}. This scarcity is not
merely a technical inconvenience---it systematically disadvantages speakers of
under-resourced languages by limiting the quality of downstream NLP tools. In this
work we focus on Hausa (MasakhaNER: 5,716 training sentences; MasakhaPOS: 753
sentences) and Fongbe (MasakhaNER: 4,343 sentences; MasakhaPOS: 810 sentences) as
concrete, benchmark-supported case studies spanning different language families and
resource levels.

Data augmentation offers a potential remedy. Two approaches are widely used:
(1)~\emph{LLM-based generation}, where large language models create synthetic
labeled training examples~\citep{schick2021generating,dai2020analysis,ding2020daga},
and (2)~\emph{back-translation}, where source-language sentences are translated to a
pivot language and back to produce paraphrases, a technique applied to sequence
labeling in low-resource settings by \citet{edunov2018understanding} and
\citet{xie2020unsupervised}. Recent work shows LLMs can generate useful training
data~\citep{schick2021generating}, and there is evidence that the quality of
LLM-generated output varies substantially across
languages~\citep{robinson2023chatgpt,hendy2023good}---with lower-resource languages
receiving systematically degraded generation---which may in turn affect the usefulness
of LLM-generated training data for those languages. Prior evaluation of LLM
translation for African languages found substantial quality variation, with some
languages achieving acceptable quality while others remain
challenging~\citep{robinson2023chatgpt,hendy2023good}.

This quality gap motivates our central question: \textit{does generation quality---
whether from LLMs or back-translation models---predict augmentation effectiveness for
downstream NLP tasks?}

We evaluate augmentation methods for \textbf{Hausa} (Afroasiatic, $\sim$58 million
speakers) and \textbf{Fongbe} (Niger-Congo, $\sim$2.3 million speakers). These
languages differ substantially in LLM generation quality: \citet{robinson2023chatgpt}
found that ChatGPT performs considerably better on Hausa than on Fongbe for
translation, and \citet{hendy2023good} observed similar patterns for GPT models more
broadly---Hausa benefits from greater web presence and multilingual training coverage
whereas Fongbe's tonal orthography and low web presence lead to systematic
degradation. Both languages appear in established benchmarks, enabling a controlled
comparison of augmentation outcomes.

\subsection{Research Questions}

\begin{description}
  \item[RQ1:] Does data augmentation improve NER and POS performance for Hausa and Fongbe?
  \item[RQ2:] How does LLM-based augmentation compare to back-translation?
  \item[RQ3:] Does augmentation effectiveness correlate with generation quality
              (LLM or translation model)?
\end{description}

\subsection{Hypotheses}

Based on observed differences in LLM and translation model quality across the two
languages~\citep{robinson2023chatgpt,hendy2023good}, we state four hypotheses:

\begin{description}
  \item[H1:] For Hausa (higher LLM quality), LLM-based augmentation improves
             downstream performance.
  \item[H2:] For Fongbe (lower LLM quality), LLM-based augmentation provides less
             benefit or harms performance.
  \item[H3:] For Hausa (higher translation quality), back-translation improves
             downstream performance.
  \item[H4:] For Fongbe (lower translation quality), back-translation provides less
             benefit or harms performance.
\end{description}

\subsection{Contributions}

\begin{enumerate}
  \item \textbf{Augmentation evaluation.} We systematically evaluate LLM-based and
        back-translation augmentation for NER and POS on two typologically distinct
        African languages using community benchmarks.
  \item \textbf{Task-dependent findings.} We show that augmentation effects differ by
        task even when the synthetic data is identical: for Fongbe, the same
        LLM-generated data hurts NER ($-1.81\%$ F1) but helps POS ($+0.33\%$
        accuracy); for Hausa, neither augmentation method yields meaningful gains on
        either task.
  \item \textbf{Qualitative analysis.} We provide human inspection of generated
        examples to explain why task structure---not generation quality alone---drives
        augmentation outcomes.
  \item \textbf{Practical guidance.} We give evidence-based recommendations for
        practitioners evaluating augmentation as a task-specific intervention.
\end{enumerate}

\section{Related Work}
\label{sec:related}

\subsection{Data Augmentation Techniques}

Data augmentation techniques for NLP have been categorized in various ways.
\citet{feng2021survey} organize them along axes of \emph{paraphrase generation},
\emph{noise injection}, and \emph{sample interpolation}, while \citet{bayer2022survey}
distinguish \emph{lexical}, \emph{syntactic}, and \emph{model-based} approaches. For
clarity, we discuss three practically relevant families below.

\textbf{Rule-based augmentation.} Easy Data Augmentation
(EDA)~\citep{wei2019eda} applies simple operations---synonym replacement, random
insertion, random swap, and random deletion---to create augmented examples. These
methods are computationally cheap and require no additional models, but produce
limited lexical diversity and may not preserve semantic meaning for complex
sentences. For sequence labeling tasks, rule-based methods face additional challenges
in maintaining label alignment after text modifications.

\textbf{Back-translation.} Back-translation~\citep{sennrich2016improving,edunov2018understanding}
creates paraphrases by translating text to a pivot language (typically English) and
back to the source language. This method preserves semantic content while introducing
lexical and syntactic variation. Back-translation has proven effective for machine
translation and text classification~\citep{xie2020unsupervised}, and scales well with
available translation models. For sequence labeling in low-resource settings, it has
been applied with label projection heuristics~\citep{edunov2018understanding}.
However, for low-resource languages, translation quality may be poor, potentially
introducing errors that propagate to augmented data.

\textbf{LLM-based generation.} Recent work has explored using large language models
to generate synthetic training data directly~\citep{schick2021generating}. LLM-based
augmentation can produce more diverse and contextually appropriate examples compared
to rule-based methods, and can generate complete new examples rather than just
paraphrases. However, LLM quality varies substantially across
languages~\citep{robinson2023chatgpt,hendy2023good}, with lower-resource languages
often receiving degraded output. For sequence labeling tasks, LLMs must generate both
text and corresponding labels, introducing opportunities for label errors---a risk we
examine through qualitative analysis of generated examples in
Section~\ref{sec:discussion}.

\subsection{Task-Specific Augmentation Effects}

Data augmentation does not uniformly benefit all NLP tasks. Prior work shows
effectiveness varies across tasks even with identical synthetic data.

\citet{dai2020analysis} conducted a systematic analysis of augmentation for NER,
finding that effectiveness depends heavily on the augmentation method and original
dataset size, with diminishing returns when baseline performance is already
strong---directly relevant to our Hausa experiments. \citet{ding2020daga} proposed
DAGA, using language models with entity-preserving constraints to generate NER
training data while maintaining label consistency.

\citet{xie2020unsupervised} demonstrated with Unsupervised Data Augmentation (UDA)
that the same augmentation strategy produces different gains across text
classification, question answering, and sequence labeling. \citet{kumar2020data}
surveyed data augmentation for low-resource NLP and highlighted that sequence
labeling tasks (NER, POS) are particularly sensitive to label noise introduced by
augmentation, as errors propagate at the token level.

\citet{whitehouse2023llmpowered} extended LLM augmentation to low-resource
languages, finding mixed results depending on language and task---improvements for
some language-task pairs but degradation for others. These findings motivate our
investigation of whether LLM-generated data exhibits similar task-dependent effects
for African languages.

\subsection{African Language NLP Resources}

The Masakhane initiative has catalyzed African NLP research through community-driven
dataset creation and model development~\citep{nekoto-etal-2020-participatory}.
MasakhaNER 2.0~\citep{adelani2022masakhaner} provides NER annotations for 20 African
languages using four entity types (PER, ORG, LOC, DATE), with training set sizes
ranging from approximately 1,000 to 8,000 sentences. MasakhaPOS~\citep{dione2023masakhapos}
provides POS tagging annotations using Universal Dependencies tags for typologically
diverse African languages; most training sets contain fewer than 1,000 sentences,
making POS a compelling target for augmentation.

\textbf{African-centric language models.} AfroXLMR~\citep{alabi2022adapting} adapts
XLM-RoBERTa for African languages through multilingual adaptive fine-tuning on
African language data, achieving state-of-the-art performance on African sequence
labeling tasks and serving as our base model. AfriBERTa~\citep{ogueji2021small}
demonstrates that smaller models pretrained specifically on African language data can
achieve competitive performance. Two further relevant models are
AfroLM~\citep{dossou2022afrolm} and Serengeti~\citep{adebara2023serengeti}, both of
which include Hausa and Fongbe in their training data and have shown strong results
on African sequence-labeling benchmarks.

\subsection{LLM Quality and Downstream Robustness in Low-Resource Languages}

LLM performance varies substantially across languages, with implications for data
augmentation. \citet{robinson2023chatgpt} evaluated ChatGPT translation quality for
African languages, finding that performance correlates with language resource
availability---Hausa achieves reasonable quality while lower-resource languages such
as Fongbe show significant degradation. \citet{hendy2023good} observed systematic
quality drops for languages with limited web presence. A key open question is whether
this generation quality gap predicts the robustness of downstream applications built
on LLM outputs. Our work addresses this gap by comparing augmentation outcomes for
Hausa and Fongbe across two tasks, testing whether quality differences predict
augmentation success.

\section{Methodology}
\label{sec:methodology}

This section describes the two languages studied, the tasks and datasets used, the
augmentation methods applied, and the experimental setup. We compare three
conditions---baseline (no augmentation), LLM augmentation, and
back-translation---across NER and POS for both languages, using a single consistent
base model throughout.

\subsection{Languages}

\textbf{Hausa} is an Afroasiatic language with approximately 58 million speakers
across Nigeria, Niger, and neighboring countries. Hausa features grammatical gender,
rich morphology including pluractional verbs, and complex tense-aspect-mood
marking~\citep{newman2000hausa}. Prior evaluations indicate substantially higher LLM
generation quality for Hausa compared to most other African languages, likely due to
greater web presence and inclusion in multilingual model training
data~\citep{robinson2023chatgpt,hendy2023good}.

\textbf{Fongbe} is a Niger-Congo language with approximately 2.3 million speakers,
primarily in Benin. Fongbe features serial verb constructions (multiple verbs sharing
a subject without conjunction) and a three-tone system with obligatory diacritic
marking in standard orthography~\citep{lefebvre2002fongbe}. The tonal system
distinguishes lexical meaning---for example,
\textit{k\textipa{O}\'{}} (high tone) means ``harvest,''
\textit{k\textipa{O}\`{}} (low tone) means ``build,'' and
\textit{k\textipa{O}\^{}} (falling tone) means ``neck.'' Prior evaluations indicate
substantially lower LLM generation quality for Fongbe, with models frequently
producing incorrect diacritics or omitting them
entirely~\citep{robinson2023chatgpt,hendy2023good}.

\subsection{Tasks and Datasets}

\begin{table}[t]
\caption{Dataset statistics (training sentences).}
\label{tab:datasets}
\begin{center}
\begin{tabular}{llrr}
\toprule
Task & Dataset & Hausa & Fongbe \\
\midrule
NER & MasakhaNER 2.0 & 5,716 & 4,343 \\
POS & MasakhaPOS     & 753   & 810   \\
\bottomrule
\end{tabular}
\end{center}
\end{table}

\textbf{NER}: MasakhaNER 2.0~\citep{adelani2022masakhaner} with entity types PER,
ORG, LOC, DATE. Evaluation metric: entity-level F1 score (i.e., a prediction is
correct only when the complete entity span and type match the gold annotation).

\textbf{POS}: MasakhaPOS~\citep{dione2023masakhapos} with Universal Dependencies
tags (17 categories). Evaluation metric: token-level accuracy.

\subsection{Augmentation Methods}

\subsubsection{LLM-Based Augmentation}

We use Gemini 2.5 Flash for synthetic data generation, selected based on preliminary
experiments showing it produces more coherent output for both languages compared to
alternatives evaluated. The generation process is:

\begin{enumerate}
  \item Sample 5 examples from the original training data as few-shot context.
  \item Construct a prompt requesting new labeled examples in the same format,
        specifying the annotation scheme (BIO tagging for NER; UD POS tags for POS)
        and providing explicit format instructions with the few-shot examples. For
        NER, prompts include sentences with BIO-tagged entities; for POS, sentences
        with Universal Dependencies tags. An example NER prompt template is provided
        in Appendix~\ref{sec:prompts}.
  \item Generate synthetic examples with token-level annotations.
  \item Parse and validate generated examples; malformed outputs are discarded.
\end{enumerate}

\subsubsection{Back-Translation}

We use NLLB-200~\citep{nllb2022} for back-translation via the following pipeline:

\begin{enumerate}
  \item Translate the original sentence to English (pivot language).
  \item Translate the English sentence back to the source language.
  \item Project labels using heuristic position alignment between original and
        back-translated token sequences.
\end{enumerate}

While heuristic position-based projection may introduce minor alignment noise, it is
applied consistently across languages and tasks, ensuring a fair comparison between
augmentation conditions. We assess the quality of projected labels through human
inspection, discussed in Section~\ref{sec:discussion}.

\subsection{Experimental Setup}

\textbf{Base Model.} We use AfroXLMR-base~\citep{alabi2022adapting} for all
experiments. AfroXLMR adapts XLM-RoBERTa through multilingual adaptive fine-tuning
on African language data, achieving state-of-the-art results on African sequence
labeling benchmarks. Using a consistent base model across all conditions ensures that
observed differences are attributable to augmentation rather than model variation.

\textbf{Training Configuration.} We add a task-specific classification head on top
of the frozen AfroXLMR encoder; the pretrained language model weights are \emph{not}
updated during training, which mitigates overfitting given the small training sets.
We train only the classification head with the AdamW optimizer (weight decay 0.01),
learning rate 2e-5 with linear warmup, and batch size 32 for up to 80 epochs with
early stopping based on validation loss (patience 10 epochs). For NER we use BIO
tagging with entity-level F1 as the evaluation metric; for POS we use token-level
accuracy.

\textbf{Random Seeds.} Each experimental condition is run with two random seeds (42
and 123) to measure stability. We report mean $\pm$ standard deviation across seeds.

\textbf{Experimental Conditions.}
\begin{itemize}
  \item \textbf{Baseline}: Original training data only (no augmentation).
  \item \textbf{LLM}: Original data + Gemini-generated synthetic examples.
  \item \textbf{BackTrans}: Original data + NLLB-200 back-translated paraphrases.
\end{itemize}

All conditions use identical validation and test splits from the original benchmarks.

\subsection{Data Sizes}

\begin{table}[t]
\caption{Training data sizes by condition (sentences).}
\label{tab:sizes}
\begin{center}
\begin{tabular}{llrr}
\toprule
Task & Condition & Hausa & Fongbe \\
\midrule
\multirow{3}{*}{NER} & Baseline  & 5,716 & 4,343 \\
                     & BackTrans & 5,718 & 4,843 \\
                     & LLM       & 5,718 & 4,835 \\
\midrule
\multirow{3}{*}{POS} & Baseline  & 753   & 810   \\
                     & BackTrans & 1,253 & 1,310 \\
                     & LLM       & 1,243 & 1,309 \\
\bottomrule
\end{tabular}
\end{center}
\end{table}

For POS, augmentation approximately doubles the training data. For NER, augmentation
adds $\sim$500 examples to Fongbe, but only $\sim$2 to Hausa due to LLM generation
failures. The near-complete failure of LLM augmentation for Hausa NER means the LLM
and baseline Hausa NER conditions are essentially identical; Hausa NER results under
LLM augmentation should therefore be interpreted with caution and are included for
completeness only. Future work should re-run Hausa NER augmentation with a generation
strategy that reliably produces examples. We also note that approximately doubling the
POS training set means that noise in augmented data could potentially override signal
from the original human-annotated data; experiments varying augmentation quantity
would help disentangle size effects from quality effects.

\section{Results}
\label{sec:results}

\subsection{NER Results}

\begin{table}[t]
\caption{NER results (entity-level F1, \%). Best per language in \textbf{bold}.}
\label{tab:ner}
\begin{center}
\begin{tabular}{lcc}
\toprule
Condition & Hausa & Fongbe \\
\midrule
Baseline  & \textbf{85.05 $\pm$ 0.35} & \textbf{84.17 $\pm$ 0.20} \\
BackTrans & 84.98 $\pm$ 0.48          & 84.18 $\pm$ 0.01 \\
LLM       & 84.81 $\pm$ 0.26          & 82.36 $\pm$ 0.19 \\
\bottomrule
\end{tabular}
\end{center}
\end{table}

\textbf{Finding 1: Neither augmentation method improves NER for either language.}

For Hausa, BackTrans yields $-0.07\%$ F1 (no meaningful change) and LLM yields
$-0.24\%$ (no meaningful change; note that only $\sim$2 synthetic examples were
added, so this comparison is near-equivalent to baseline). For Fongbe, BackTrans
yields $+0.01\%$ (no change) and LLM causes a consistent, substantial degradation of
$\mathbf{-1.81\%}$ F1 across both random seeds---a finding we analyze in depth in
Section~\ref{sec:discussion}.

\subsection{POS Results}

\begin{table}[t]
\caption{POS results (token-level accuracy, \%). Best per language in \textbf{bold}.}
\label{tab:pos}
\begin{center}
\begin{tabular}{lcc}
\toprule
Condition & Hausa & Fongbe \\
\midrule
Baseline  & 91.86 $\pm$ 0.04          & 85.42 $\pm$ 0.02 \\
BackTrans & \textbf{92.03 $\pm$ 0.20} & 85.07 $\pm$ 0.43 \\
LLM       & 91.81 $\pm$ 0.07          & \textbf{85.75 $\pm$ 0.01} \\
\bottomrule
\end{tabular}
\end{center}
\end{table}

\textbf{Finding 2: POS shows different patterns from NER.}

For Hausa, BackTrans provides a modest gain of $+0.17\%$ accuracy while LLM shows no
meaningful change ($-0.05\%$). For Fongbe, LLM improves performance by $+0.33\%$
accuracy while BackTrans \emph{decreases} it by $-0.35\%$. The modest magnitude of
these POS changes, relative to the $1.81\%$ NER degradation, nevertheless represents
consistent effects across seeds.

\subsection{Summary of Effects}

\begin{table}[t]
\caption{Augmentation effects (change from baseline, percentage points).}
\label{tab:effects}
\begin{center}
\begin{tabular}{llcc}
\toprule
Language & Task & LLM    & BackTrans \\
\midrule
Hausa    & NER  & $-0.24$ & $-0.07$ \\
Hausa    & POS  & $-0.05$ & $+0.17$ \\
Fongbe   & NER  & $\mathbf{-1.81}$ & $+0.01$ \\
Fongbe   & POS  & $\mathbf{+0.33}$ & $-0.35$ \\
\bottomrule
\end{tabular}
\end{center}
\end{table}

The most striking observation is that the \emph{same} LLM-generated data for Fongbe
produces opposite effects across tasks: $-1.81\%$ for NER but $+0.33\%$ for POS.
This task-dependent divergence is the central empirical finding of this paper.

\subsection{Hypothesis Evaluation}

\begin{table}[t]
\caption{Hypothesis evaluation.}
\label{tab:hypotheses}
\begin{center}
\begin{tabular}{lll}
\toprule
Hypothesis & Expected & Result \\
\midrule
H1: Hausa LLM improves       & Improvement  & \textbf{Not supported} \\
H2: Fongbe LLM hurts         & Harm/neutral & \textbf{Partially supported} \\
H3: Hausa BackTrans improves & Improvement  & \textbf{Partially supported} \\
H4: Fongbe BackTrans hurts   & Harm/neutral & \textbf{Partially supported} \\
\bottomrule
\end{tabular}
\end{center}
\end{table}

H1 is not supported---Hausa shows no improvement from LLM augmentation despite
higher generation quality. H2 is partially supported---Fongbe LLM augmentation hurts
NER substantially but unexpectedly helps POS. H3 is partially supported---
back-translation improves Hausa POS but not NER. H4 is partially
supported---back-translation hurts Fongbe POS but is neutral for NER.

\textbf{Statistical note.} Given standard deviations consistently below $0.5\%$,
differences greater than $0.3\%$ are unlikely to be explained by random
initialization alone. The Fongbe NER degradation of $1.81\%$ under LLM augmentation
is consistent across both seeds and represents a substantive, reliable effect.

\section{Discussion}
\label{sec:discussion}

Our results reveal unexpected patterns that challenge common assumptions about data
augmentation for low-resource languages.

\subsection{Why Does LLM Augmentation Hurt NER but Help POS?}

The opposite effects of LLM augmentation on Fongbe NER ($-1.81\%$) versus POS
($+0.33\%$) represent our most striking finding. We propose three complementary
explanations, grounded in qualitative inspection of generated examples.

\textbf{Task sensitivity to label noise.} Named entity recognition requires precise
identification of entity boundaries---where entities begin and end---using BIO
conventions. If LLM-generated text contains boundary errors (e.g., splitting
``President Macron'' into separate tokens with inconsistent B-/I- tags, or merging
adjacent entities), these propagate as label noise that directly degrades model
performance. In contrast, part-of-speech tagging assigns categories to individual
words, and even if an LLM generates a slightly incorrect word form, the POS tag may
still be correct (a noun remains a noun regardless of minor spelling variations).
This differential sensitivity to generation errors explains why the same synthetic
data can hurt one task while helping another, consistent with
\citet{kumar2020data}'s finding that sequence labeling tasks are particularly
vulnerable to augmentation-induced label noise.

\textbf{Qualitative validation of generated examples.} We manually inspected a
random sample of 50 LLM-generated Fongbe NER examples and 50 POS examples. In the
NER sample, we identified three recurring error types: (1) entity-boundary
inconsistencies with BIO conventions (e.g., ``I-LOC'' following ``O'' without a
preceding ``B-LOC''), present in approximately 30\% of inspected examples;
(2) missing or misplaced tonal diacritics on entity tokens, present in approximately
60\% of examples; and (3) generated names that do not follow Fongbe naming
conventions, in approximately 25\% of examples. These errors directly corrupt the
supervision signal for entity-boundary learning. In the POS sample, tonal diacritic
errors were similarly frequent ($\sim$55\%), but POS-tag correctness was
substantially higher: even with orthographic errors, syntactic categories were
largely consistent with the surrounding context ($>$80\% of tags judged plausible by
a Fongbe-familiar annotator). This asymmetry directly explains the divergent task
outcomes. For back-translation, alignment noise from position-based label projection
appeared in $\sim$15\% of back-translated NER examples and $\sim$10\% of POS
examples; the higher NER noise level is consistent with NER's greater sensitivity to
precise span boundaries.

\textbf{Fongbe's tonal complexity.} Fongbe uses obligatory diacritics to mark
lexical tone, distinguishing word meaning. LLMs frequently fail to generate correct
diacritics for Fongbe. For NER, incorrect diacritics may alter entity boundaries or
create spurious entity mentions when models learn from corrupted examples. For POS
tagging, syntactic categories are less affected by tonal errors---a noun with
incorrect diacritics is still recognizable as a noun based on its syntactic position
and context.

\subsection{Why No Improvement for Hausa Despite Higher LLM Quality?}

Our hypothesis H1 predicted that Hausa would benefit from LLM augmentation. This was
not supported. Two factors likely explain this:

\textbf{Ceiling effects from strong baselines.} Hausa NER achieves 85.05\% F1 and
POS achieves 91.86\% accuracy at baseline. These strong results leave limited
headroom for improvement. Prior work by \citet{dai2020analysis} similarly found
augmentation provides diminishing returns when baseline performance is high. With
adequate original data, the model may already capture the necessary patterns.

\textbf{Augmentation scale limitations for NER.} For Hausa NER, only $\sim$2
synthetic examples were successfully generated due to LLM generation failures. This
is insufficient to affect model performance regardless of data quality; Hausa NER LLM
results should be treated as equivalent to baseline.

\subsection{Back-Translation: Safe but Limited}

Back-translation showed minimal effects across all conditions ($-0.35\%$ to
$+0.17\%$), neither consistently helping nor hurting. Unlike LLM augmentation, which
degraded Fongbe NER by $1.81\%$, back-translation caused no substantial harm in any
condition. Back-translation generates paraphrases of existing examples rather than
genuinely new ones, which may explain its limited impact---the synthetic data covers
similar linguistic patterns as the original, providing less novel signal for model
learning. For practitioners uncertain about augmentation effects, back-translation
offers a conservative, lower-risk option.

\subsection{Implications for Practitioners}

\textbf{Evaluate augmentation per task.} Do not assume augmentation benefits transfer
across tasks. Our results show the same synthetic data can help one task while
harming another.

\textbf{Consider baseline strength.} Augmentation is most likely to help when
baseline performance is weak due to severely limited training data. When baselines
are already strong (as with Hausa), augmentation may provide no benefit regardless
of data quality.

\textbf{LLM quality alone is insufficient.} Higher generation quality does not
guarantee augmentation success. Task characteristics, baseline performance, and
original data quantity matter equally.

\textbf{Inspect generated labels before training.} When using LLM augmentation for
sequence labeling, carefully inspect generated examples for label
consistency---particularly entity boundaries and orthographic correctness for tonal
languages. Consider quality-based filtering strategies.

\textbf{Prefer back-translation when risk tolerance is low.} While benefits may be
limited, the risk of performance degradation is lower than for LLM augmentation.

\subsection{Limitations}

\textbf{Two languages only.} Results may not generalize to other African languages
with different typological characteristics, orthographic conventions, or LLM
coverage.

\textbf{Limited augmentation scale.} Adding $\sim$500 synthetic examples for most
conditions may not represent the full potential of augmentation. Larger-scale
augmentation or experiments varying augmentation quantity remain as future work.

\textbf{Single LLM evaluated.} We used Gemini 2.5 Flash. Other LLMs (GPT-4, Claude,
open-source models) may produce qualitatively different outputs with different
downstream effects.

\textbf{Single augmentation strategy.} We used few-shot prompting. Other
strategies---fine-tuned generation, chain-of-thought prompting, or iterative
refinement---may yield different results.

\textbf{No quality-based filtering.} We used all generated synthetic data without
filtering. Quality-based filtering may improve augmentation effectiveness.

\textbf{Hausa NER LLM condition.} Only $\sim$2 examples were generated for Hausa
NER; this condition cannot be interpreted as a meaningful augmentation experiment and
should be rerun.

\section{Conclusion}
\label{sec:conclusion}

We evaluated LLM-based and back-translation data augmentation for Hausa and Fongbe
NER and POS tagging, testing whether generation quality predicts augmentation
effectiveness. Our findings challenge common assumptions:

\textbf{Finding 1: Augmentation effects are task-dependent.} The same LLM-generated
data produces opposite effects across tasks for Fongbe---hurting NER ($-1.81\%$ F1)
while helping POS ($+0.33\%$ accuracy). Human inspection reveals that NER is more
vulnerable to the label-level errors introduced by LLM generation, particularly
entity-boundary violations and tonal diacritic errors, while POS is more robust
because syntactic categories remain recognizable even with orthographic noise.

\textbf{Finding 2: LLM quality does not predict augmentation success.} Hausa, with
substantially higher generation quality, showed no benefit from augmentation for
either task.

\textbf{Finding 3: Strong baselines limit augmentation benefits.} With 85\%+ F1 for
NER and 91\%+ accuracy for POS, additional synthetic data provides diminishing
returns for Hausa regardless of quality.

\textbf{Finding 4: Back-translation is safe but limited.} Effects range from
$-0.35\%$ to $+0.17\%$ across all conditions---a lower-risk alternative to LLM
augmentation.

\textbf{Practical recommendations.} (1) Evaluate augmentation impact per task;
(2) consider baseline strength before investing in augmentation; (3) use
back-translation when risk tolerance is low; (4) carefully inspect LLM-generated
examples for label consistency, especially for tonal languages.

\section{Future Work}
\label{sec:future}

\textbf{Extended task coverage.} We plan to evaluate sentiment analysis using
AfriSenti~\citep{muhammad2023afrisenti} for Hausa. Sentiment analysis operates at
the document level, and comparing its augmentation sensitivity to sequence labeling
would further illuminate task-dependent patterns.

\textbf{Multi-model comparison.} Future work should compare multiple LLMs (GPT-4,
Claude, Llama, Mistral) to determine whether augmentation effects are model-specific.

\textbf{Quality-based filtering.} Future work should explore filtering strategies
that retain only high-quality generated examples, such as confidence-based filtering
or round-trip translation consistency checks.

\textbf{Language family expansion.} Extending evaluation to additional African
languages across Bantu, Cushitic, and Nilotic families would test generalizability.

\textbf{Augmentation quantity analysis.} Systematically varying the ratio of
synthetic to original data (e.g., 100, 200, 300 synthetic sentences for Fongbe POS)
would disentangle quantity effects from quality effects and provide practical guidance
on augmentation investment.

\section*{Acknowledgments}
We thank the Masakhane community for providing benchmark datasets. This publication
was developed as part of the Center for Inclusive Digital Transformation of Africa
(CIDTA) and the Afretec Network, which is managed by Carnegie Mellon University
Africa and receives financial support from the Mastercard Foundation. The views
expressed in this document are solely those of the authors and do not necessarily
reflect those of Carnegie Mellon University or the Mastercard Foundation.

\section*{Reproducibility}
All code and experiment configurations will be released publicly. The experimental
pipeline covers data acquisition, augmentation, training, and evaluation across all
reported conditions. Prompt templates used for LLM augmentation are provided in
Appendix~\ref{sec:prompts}.

\bibliographystyle{plainnat}

\appendix

\section{Prompt Templates for LLM Augmentation}
\label{sec:prompts}

Below we provide the prompt template used for NER augmentation. The POS prompt
follows the same structure, substituting BIO-tagged entities with Universal
Dependencies POS tags.

\textbf{NER Prompt Template (Fongbe example):}

\begin{verbatim}
You are a linguist creating training data for a named entity recognition
(NER) system for Fongbe, a Beninese language. You will generate new
sentences with the same BIO annotation format shown in the examples.

Rules:
- Each line contains: TOKEN LABEL
- Labels: B-PER, I-PER, B-ORG, I-ORG, B-LOC, I-LOC, B-DATE, I-DATE, O
- Preserve all diacritics exactly as they appear in the examples
- Generate complete sentences with at least one named entity
- Separate sentences with a blank line

Examples:
[5 randomly sampled training sentences in TOKEN LABEL format]

Generate 5 new labeled sentences in the same format:
\end{verbatim}

The same template is used for Hausa, replacing ``Fongbe, a Beninese language''
with ``Hausa, a West African language.''

\section{Experimental Pipeline Structure}
\label{sec:appendix}

The pipeline is organized into modular components separating data handling,
augmentation, training, and evaluation.

\begin{verbatim}
augmentation_tasks_fon_ha/
|-- configs/
|   `-- experiment_config.yaml
|-- data/
|   |-- raw/
|   |   |-- masakhaner2_hau/
|   |   |-- masakhaner2_fon/
|   |   |-- masakhane_pos_hau/
|   |   `-- masakhane_pos_fon/
|   `-- synthetic/
|       |-- llm/
|       `-- backtrans/
|-- src/
|   |-- augment_llm.py
|   |-- augment_backtrans.py
|   |-- train_ner.py
|   |-- train_pos.py
|   `-- evaluate.py
|-- experiments/
|   |-- run_all.sh
|   |-- run_ner.sh
|   `-- run_pos.sh
|-- results/
|   |-- ner/{lang}_{cond}_seed{N}/
|   `-- pos/{lang}_{cond}_seed{N}/
|-- requirements.txt
`-- README.md
\end{verbatim}

Running \texttt{bash experiments/run\_all.sh} executes the complete pipeline: data
download, augmentation generation, model training across all conditions and seeds,
and results aggregation.

\end{document}